\documentclass[manuscript]{acmart}
\AtBeginDocument{%
  \providecommand\BibTeX{{%
    \normalfont B\kern-0.5em{\scshape i\kern-0.25em b}\kern-0.8em\TeX}}}

\copyrightyear{2023}
\acmYear{2023}
\setcopyright{licensedusgovmixed}\acmConference[KDD '23]{Proceedings of the 29th ACM SIGKDD Conference on Knowledge Discovery and Data Mining}{August 6--10, 2023}{Long Beach, CA, USA}
\acmBooktitle{Proceedings of the 29th ACM SIGKDD Conference on Knowledge Discovery and Data Mining (KDD '23), August 6--10, 2023, Long Beach, CA, USA}
\acmPrice{15.00}
\acmDOI{10.1145/3580305.3599819} \acmISBN{979-8-4007-0103-0/23/08}

\def\L\mathcal{L}

\def\V{\mathbf{V}}

\def\L{\mathcal{L}}

\def\I{\mathbf{I}}

\def\I{\mathbf{I}}

\def\q{\mathbf{q}} 
\def\f{\mathbf{f}}

\def\0{\mathbf{0}}
\def\I{\mathbf{I}}

\def\a{\mathbf{a}}

\def\L{\mathcal{L}}

\usepackage{hyperref}
\usepackage{url}
\usepackage{graphicx}
\usepackage{multirow}
\usepackage{booktabs}
\usepackage{ bbold }
\usepackage{caption}
\usepackage{subcaption}
\usepackage{balance} 

\begin{document}

\title[Expert Knowledge-Aware Medical Image Difference VQA]{Expert Knowledge-Aware Image Difference Graph Representation Learning for Difference-Aware Medical Visual Question Answering}

\author{Xinyue Hu}
\affiliation{
    \institution{The University of Texas at Arlington}
    \streetaddress{701 S Nedderman Dr}
    \city{Arlington}
    \state{Texas}
    \country{USA}
    \postcode{76019}
}
\email{xxh4034@mavs.uta.edu}

\author{Lin Gu}
\affiliation{
    \institution{RIKEN}
    \city{Tokyo}
    \country{Japan}
}
\affiliation{
    \institution{The University of Tokyo}
    \city{Tokyo}
    \country{Japan}
}
\email{lin.gu@riken.jp}

\author{Qiyuan An}
\affiliation{
    \institution{The University of Texas at Arlington}
    \streetaddress{701 S Nedderman Dr}
    \city{Arlington}
    \state{Texas}
    \country{USA}
    \postcode{76019}
}
\email{qxa5560@mavs.uta.edu}

\author{Mengliang Zhang}
\affiliation{
    \institution{The University of Texas at Arlington}
    \streetaddress{701 S Nedderman Dr}
    \city{Arlington}
    \state{Texas}
    \country{USA}
    \postcode{76019}
}
\email{mxz3935@mavs.uta.edu}

\author{Liangchen Liu}
\affiliation{
    \institution{National Institutes of Health Clinical Center}
    \streetaddress{10 Center Dr}
    \city{Bethesda}
    \state{Maryland}
    \country{USA}
    \postcode{20892}
}
\email{liangchen.liu@nih.gov}

\author{Kazuma Kobayashi}
\affiliation{
    \institution{National Cancer Center Research Institute}
    \city{Tokyo}
    \country{Japan}
}
\email{kazumkob@ncc.go.jp}

\author{Tatsuya Harada}
\affiliation{
    \institution{The University of Tokyo}
    \city{Tokyo}
    \country{Japan}
}
\affiliation{
    \institution{RIKEN}
    \city{Tokyo}
    \country{Japan}
}
\email{harada@mi.t.u-tokyo.ac.jp}

\author{Ronald M. Summers}
\affiliation{
    \institution{National Institutes of Health Clinical Center}
    \streetaddress{10 Center Dr}
    \city{Bethesda}
    \state{Maryland}
    \country{USA}
    \postcode{20892}
}
\email{rsummers@mail.cc.nih.gov}

\author{Yingying Zhu}
\authornote{Corresponding author.}
\affiliation{
    \institution{The University of Texas at Arlington}
    \streetaddress{701 S Nedderman Dr}
    \city{Arlington}
    \state{Texas}
    \country{USA}
    \postcode{76019}
}
\email{yingying.zhu@uta.edu}

\renewcommand{\shortauthors}{Xinyue Hu et al.}

\begin{abstract}
To contribute to automating the medical vision-language model, we propose a novel Chest-Xray Difference Visual Question Answering (VQA) task. Given a pair of main and reference images, this task attempts to answer several questions on both diseases and, more importantly, the differences between them. 
This is consistent with the radiologist's diagnosis practice that compares the current image with the reference before concluding the report. We collect a new dataset, namely  MIMIC-Diff-VQA, including 700,703 QA pairs from 164,324 pairs of main and reference images.  Compared to existing medical VQA datasets, our questions are tailored to the Assessment-Diagnosis-Intervention-Evaluation treatment procedure used by clinical professionals.
Meanwhile, we also propose a novel expert knowledge-aware graph representation learning model to address this task. 
The proposed baseline model leverages expert knowledge such as anatomical structure prior, semantic, and spatial knowledge to construct a multi-relationship graph, representing the image differences between two images for the image difference VQA task. 
The dataset and code can be found at \url{https://github.com/Holipori/MIMIC-Diff-VQA}. 
We believe this work would further push forward the medical vision language model.
\end{abstract}






\received{20 February 2007}
\received[revised]{12 March 2009}
\received[accepted]{5 June 2009}

\maketitle

\section{Introduction}
\label{intro}

%The medical informatics community
% Several recent works focus on 

% Extracting text-mined labels from clinical notes to  train deep-learning models for medical image analysis

% with several datasets:
% %Chest X-ray is one of the most commonly accessible examinations for screening and diagnosis, leading to a tremendous number of radiology images associated with free-text and reports accumulated in hospitals and released online including
%  MIMIC \cite{johnson2019mimic}, NIH14 \cite{wang2017chestx} and Chexpert \cite{irvin2019chexpert}.

\begin{figure*}[t]
\centering
\includegraphics[width=1\textwidth]{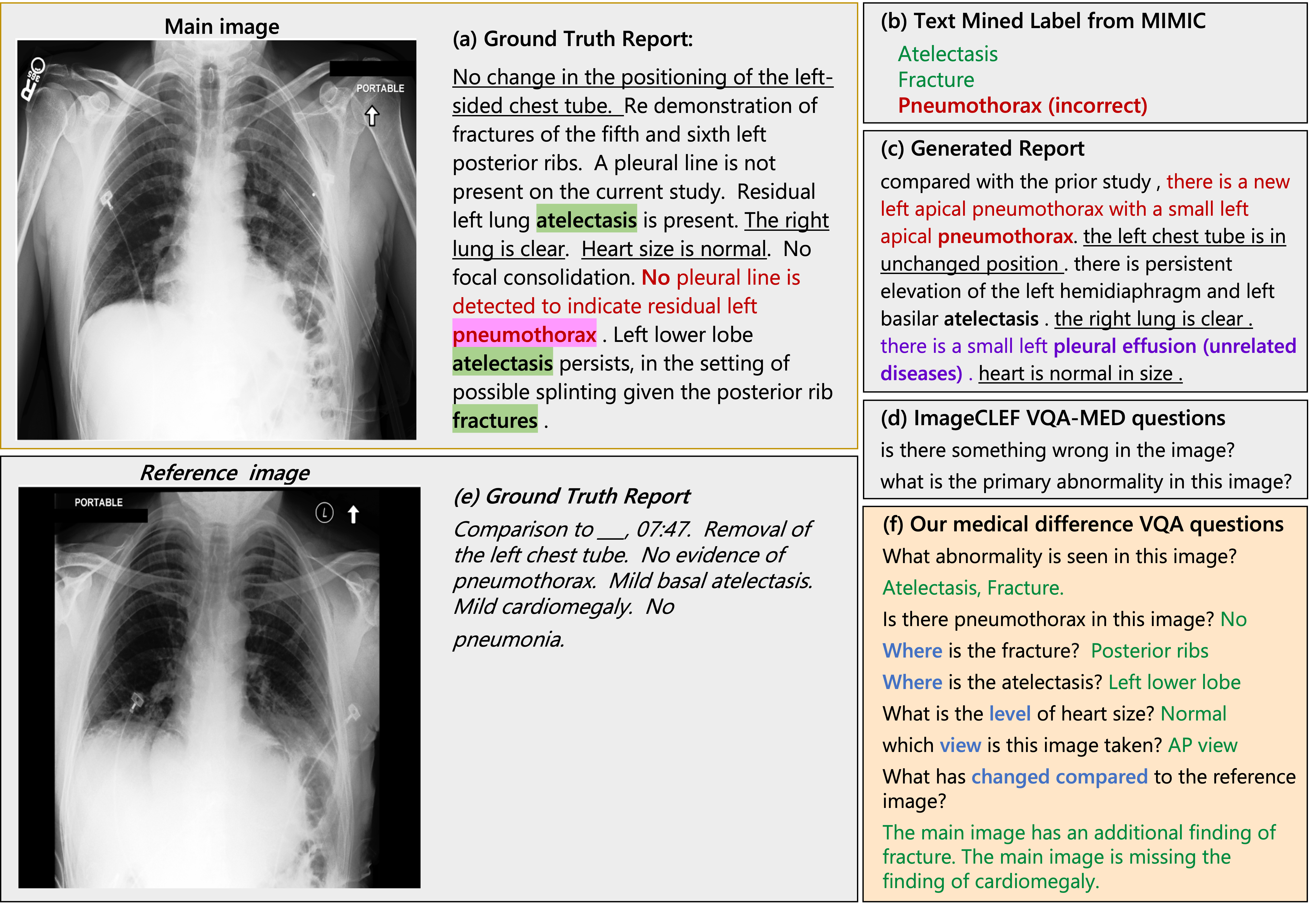}
\caption{(a) The ground truth report corresponding to the main(present) image. The red text represents labels incorrectly classified by either text mining or generated reports, while the red box marks the misclassified labels. The green box marks the correctly classified ones. The underlined text is correctly generated in the generated report. (b) The label "Pneumothorax" is incorrectly classified because there is NO evidence of pneumothorax from the chest X-ray. (c) "There is a new left apical pneumothorax" $\rightarrow$ This sentence is wrong because the evidence of pneumothorax was mostly improved after treatment. However, the vascular shadow in the left pulmonary apex is not very obvious, so it is understandable why it is misidentified as pneumothorax in the left pulmonary apex. "there is a small left pleural effusion" $\rightarrow$ It is hard for a doctor to tell if the left pleural effusion is present or not. (d) The ImageCLEF-VQA-Med questions are designed too simple. (e) The reference(past) image and clinical report. (f) Our medical difference VQA questions are designed to guide the model to focus on and localize important regions. 
}
\Description{On the left side, two chest X-ray images taken during two separate visits are displayed with their corresponding text reports. On the right side, additional text information is provided, including the traditional MIMIC mined label, a generated report, ImageCLEF-VQA-MED questions, and our medical difference VQA questions.}
\label{fig:fig1}
\end{figure*}

 The medical informatics community has been working on feeding the data-hungry deep learning algorithms by fully exploiting hospital databases with invaluable loosely labeled imaging data. Among diverse attempts, Chest X-ray datasets such as  MIMIC \cite{johnson2019mimic}, NIH14 \cite{wang2017chestx} and Chexpert \cite{irvin2019chexpert} have received particular attention.  During this arduous journey on vision-language (VL) modality, the community either mines per-image common disease label (Fig.\ref{fig:fig1}. (b)) through Natural Language Processing (NLP) or endeavors on report generation (Fig.\ref{fig:fig1}. (c) generated from \cite{emnlp2021report}) or even answer certain pre-defined questions (Fig.\ref{fig:fig1}. (d)).
Despite significant progress achieved on these tasks, the heterogeneity, systemic biases, and subjective nature of the report still pose many technical challenges. For example, the automatically mined labels from reports in Fig.\ref{fig:fig1}. (b) is problematic because the rule-based approach that was not carefully designed did not process all uncertainties and negations well \cite{johnson2019mimic-jpg}. 
Training an automatic radiology report generation system to match the report appears to avoid the inevitable bias in the standard NLP-mined thoracic pathology labels. 
However, radiologists tend to write more obvious impressions with abstract logic. For example, as shown in Fig.\ref{fig:fig1}. (a), a radiology report excludes many diseases (either commonly diagnosed or intended by the physicians) using negation expressions, e.g., no, free of, without, \textit{etc.} However, the artificial report generator could hardly guess which disease is excluded by radiologists.  
\begin{figure*}[t]
    \centering
    \includegraphics[width=0.85\textwidth]{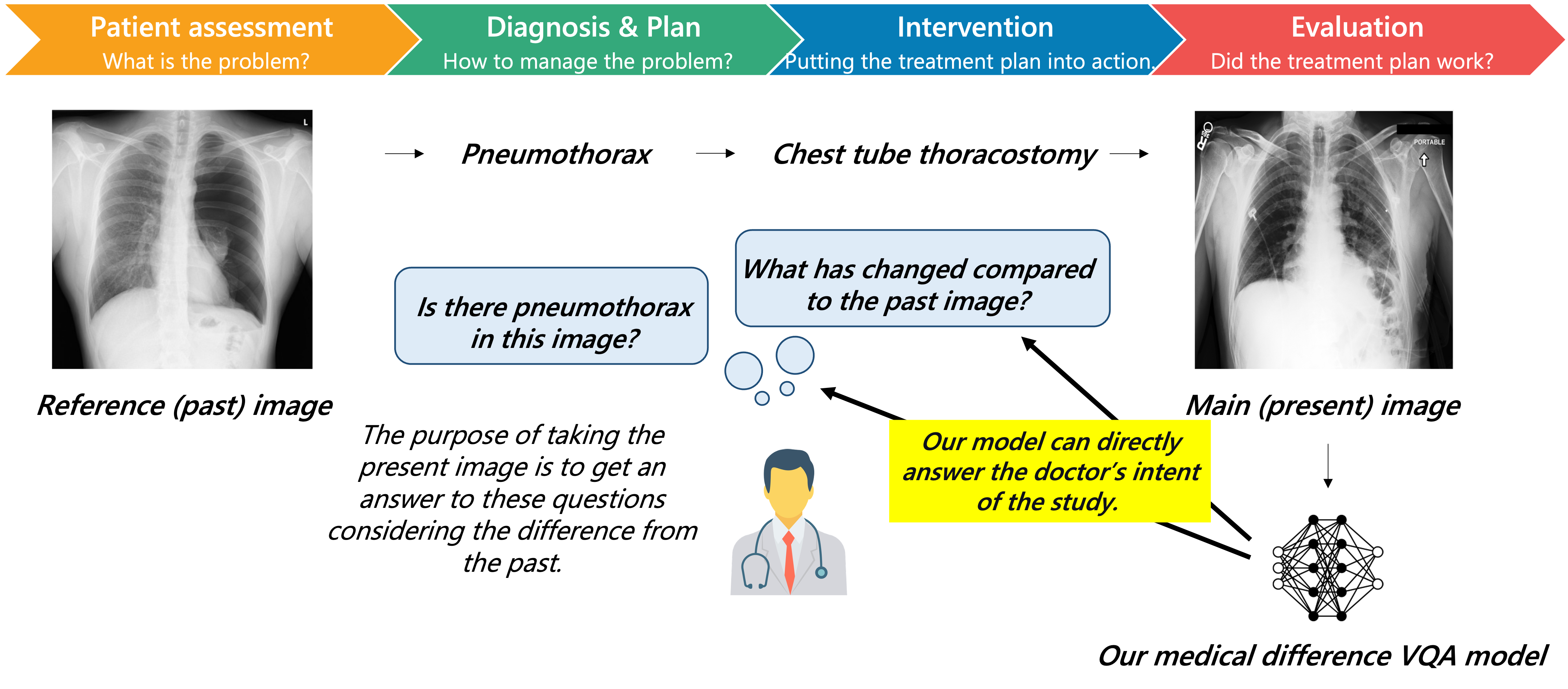}
    \caption{Clinical motivation for Image difference VQA.}
    \Description{The two chest X-ray images are displayed on the left and right sides, demonstrating the process of clinical diagnosis from left to right: patient assessment, diagnosis and plan, intervention, and evaluation. The figure provides an example of the inference from Pneumothorax to Chest tube thoracostomy. In the center, the possible questions a doctor may ask are highlighted, such as "Is there evidence of Pneumothorax in this image?" and "What changes can be seen compared to the previous image?"}
    \label{fig:motivation}
\end{figure*}
Instead of thoroughly generating all of the descriptions, VQA is more plausible as it only answers the specific question. As shown in Fig. \ref{fig:fig1}, the question could be raised strictly for "is there any pneumothorax in the image?" in the report while the answer is no doubt "No". However, the questions in the existing VQA dataset ImageCLEF-VQA-Med \cite{ImageCLEF-VQA-Med2021} concentrate on very few general ones, such as "is there something wrong in the image? what is the primary abnormality in this image?", lacking the specificity for the heterogeneity and subjective texture. It not only degrades VQA into classification but, more unexpectedly, provides little helpful information for clinics. While VQA-RAD \cite{VQA-RAD} has more heterogeneous questions covering 11 question types, its 315 images dataset is relatively too small.

%\textcolor{black}{Kobayashi San please briefly add some lines here}.

To bridge the aforementioned gap in the visual language model, we propose a novel medical image difference VQA task more consistent with radiologists' practice.
% When radiologists make a diagnosis, they usually compare the main one with a reference image to find their differences.
% Therefore, understanding the clinical meaning of what and where has changed on two images is an essential step in the arduous medical vision language journey.  
{
\color{black}When radiologists make diagnoses, they compare current and previous images of the same patients to check the disease's progress. Actual clinical practice follows a patient treatment process (assessment - diagnosis - intervention - evaluation) as shown in Fig.~\ref{fig:motivation}.
A baseline medical image is used as an assessment tool to diagnose a clinical problem, usually followed by therapeutic intervention. Then, another follow-up medical image is retaken to evaluate the effectiveness of the intervention in comparison with the past baseline.  In this framework, every medical image has its purpose of clarifying the doctor's clinical hypothesis depending on the unique clinical course (\textit{e.g.}, whether the pneumothorax is mitigated after therapeutic intervention). 
However, existing methods can not provide a straightforward answer to the clinical hypothesis since they do not compare the past and present images. 
Therefore, we present a chest X-ray image difference VQA dataset, MIMIC-Diff-VQA, to fulfill the need of the medical image difference task. Moreover, we propose a system that answers doctors' questions by comparing the current medical image (main) to a past visit medical image (reference). This allows us to build a diagnostic support system that realizes the inherently interactive nature of radiology reports in clinical practice.  
}

% Therefore, we present a chest X-ray image difference VQA dataset, MIMIC-Diff-VQA.  
MIMIC-Diff-VQA contains pairs of "main"(present) and "reference"(past) images from the same patient's radiology images at different times from MIMIC-CXR\cite{johnson2019mimic} (a large-scale public database of chest radiographs with 227,835 studies, each with a unique report and images).
The question and answer pairs are extracted from the MIMIC-CXR report for "main"  and "reference" images using an Extract-Check-Fix cycle. 
% Similar to \cite{VQA-Med,VQA-RAD,pathvqa}, we first collect sets of abnormality names and attributes. Then we extract the abnormality in the images and their corresponding attributes using regular expressions. Finally, we compare the abnormalities contained in the two images and ask questions based on the collected information. 
There are seven types of questions included in our dataset: 1. abnormality, 2. presence, 3. view, 4. location, 5. type, 6. level, and 7. difference. The MIMIC-Diff-VQA dataset comprises 700,703 QA pairs extracted from 164,324 image pairs. 
% \textcolor{black}{which has been filtered to only contains the images from the PA view and AP view, and the corresponding patient has more than one image.}
Particularly, \textit{difference} questions are pairs of inquiries that pertain to the clinical progress and changes in the "main" image as compared to the "reference" image, as shown in Fig.~\ref{fig:fig1}(e).

%Comparing the differences between two medical images are very challenging task due to the body pose, view variances and deformations as shown in Fig.~\ref{fig:fig1}. %Current state-of-art image difference model did not consider 

The current mainstream state-of-the-art image difference method only applies to synthetic images with small view variations,\cite{jhamtani2018learning,park2019robust} as shown in Fig.~\ref{fig:achitechture}.
However, real medical image difference comparing is a very challenging task. Even the images from the same patient show large variances in the orientation, scale, range, view, and nonrigid deformation, which are often more significant than the subtle differences caused by diseases as shown in Fig.~\ref{fig:achitechture}. 
Since the radiologists examine the anatomical structure to find the progression of diseases, similarly, we propose an expert knowledge-aware image difference graph representation learning model as shown in Fig.~\ref{fig:achitechture}. We extract the features from different anatomical structures (for example, left lower lung, and right upper lung) as nodes in the graph. 
%we proposed an anatomical structure-aware image difference graph model and compare the image differences in each normalized anatomical region (for example, left lower lung). Each anatomical structure is defined as a node in the graph and we use the pre-trained anatomical and disease region detection model to extract the feature of each node. 

\textcolor{black}{
Moreover, we construct three different relationships in the graph to encode expert knowledge: 1) Spatial relationship based on the spatial distance between different anatomical regions.
2) Semantic relationship based on the disease and anatomical structure relationship from knowledge graph~\cite{zhang2020radiology}.
3) Implicit relationship to model potential implicit relationship beside 1) and 2). 
The image-difference graph feature representation is constructed by simply subtracting the main image graph feature and the reference image graph feature. This graph difference feature is fed into LSTM networks with attention modules for answer generation\cite{toutanova2003feature}. 
}

% Moreover, we construct three different relationships in the graph to encode expert knowledge: 1) Spatial relationship based on the spatial distance between different anatomical regions, such as "left lower lung", "right costophrenic angle", etc. We design this graph based on the fact that radiologists prefer to determine the abnormalities based on particular anatomical structures. For example, "Minimal blunting of the left costophrenic angle also suggests a tiny left pleural effusion.";
% 2) Semantic relationship based on the disease and anatomical structure relationship from knowledge graph~\cite{zhang2020radiology}.
% We construct this graph because diseases from the same or nearby regions could affect each other's existence. For example, "the effusions remain moderate and still cause substantial bilateral areas of basilar atelectasis."; 
% 3) Implicit relationship to model potential implicit relationship beside 1) and 2). 
% The graph feature representation for each image is learned as a weighted summation of the graph feature from these three different relationships. The image-difference graph feature representation is constructed by simply subtracting the main image graph feature and the reference image graph feature. This graph difference feature is fed into LSTM networks with attention modules for answer generation\cite{toutanova2003feature}. 

\textbf{ Our contributions are summarized as:}
%\textcolor{black}{ADD how the post processing part for feature extraction here and explain why this post featue processing are important for this, and citep the ACL paper here.: HXY }
%We constructed two graphs on the main and reference images and subtracting
%Next, a difference feature is generated and is fed into the feature attention module along with the graph features to generate the final features of the images. 
%Finally, the final answer is generated by the answer generator, which consists of attention modules and LSTM networks that takes into account Part-Of-Speech~\cite{toutanova2003feature} information.
%\textcolor{black}{
%In the general image difference\cite{tu2021semantic}, their before and after image are overall similar and easy to localize the difference by directly apply the feature maps.
%However, even if two chest X-ray images are from the same patient, the images can be very different due to the different positions at the time of capture.
%Therefore, we construct each image feature with combined anatomical features and their corresponding disease features.
%The anatomical feature is constructed in a specific order of anatomical locations.
%By doing so, we ensure that the model knows what anatomical structures each feature represents.
%}

1)We propose the medical imaging difference visual question answering problem and construct the first large-scale medical image difference visual question answering dataset, MIMIC-Diff-VQA. This dataset comprises  \textcolor{black}{164,324} image pairs,  containing \textcolor{black}{700,703} question-answer pairs related to various attributes, including abnormality, presence, location, level, type, view, and difference. 

2) We propose an anatomical structure-aware image-difference model to extract the image-difference feature relevant to disease progression and interventions. We extracted features from anatomical structures and compared the changes in each structure to reduce the image differences caused by body pose, view, and nonrigid deformations of organs.  

3)  We develop a multi-relationship image-difference graph feature representation learning method to leverage the spatial relationship and semantic relationship (extracted from expert knowledge graph) to compute image-difference graph feature representation, generate answers and interpret how the answer is generated on different image regions.

%both disease findings and anatomical structures are defined as nodes. We also establish the spatial relation among these nodes.

\section{MIMIC-Diff-VQA dataset.}
We introduce our new MIMIC-Diff-VQA dataset for the medical imaging difference question-answering problem. 
{
\color{black}
The MIMIC-Diff-VQA dataset is constructed following an Extract-Check-Fix cycle to minimize errors.
% Please refer to Appendix.~\ref{sec:dataset_construct} for the details on how the dataset is constructed.
}
In MIMIC-Diff-VQA, each entry contains two different chest X-ray images from the same patient with a question-answer pair.
Our question design is extended from VQA-RAD, but with an additional "difference" question type.
Ultimately, the questions can be divided into seven types: 1) abnormality, 2) presence, 3) view, 4) location, 5) type, 6) level, and 7) difference. 
Tab.~\ref{tab:question} shows examples of the different question types.

\begin{table*}[h]
  \caption{Selected examples of the different question types. See the Appendix for the full list.}
  \label{tab:question}
  \centering
    \begin{tabular}{ll}
    \toprule
    % \hline
    Question type & Example                                             \\
    \midrule
    % \hline \\
    Abnormality   & what abnormality is seen in the left lung?          \\ 
    Presence      & is there evidence of atelectasis in this image?     \\ 
    View          & which view is this image taken?                     \\ 
    Location      & where in the image is the pleural effusion located? \\ 
    Type          & what type is the opacity?                           \\
    Level         & what level is the cardiomegaly?                     \\ 
    Difference    & what has changed compared to the reference image?   \\
    \bottomrule
    \end{tabular}
\end{table*}

\begin{figure}[h]
    \centering
    \includegraphics[width=0.4\textwidth]{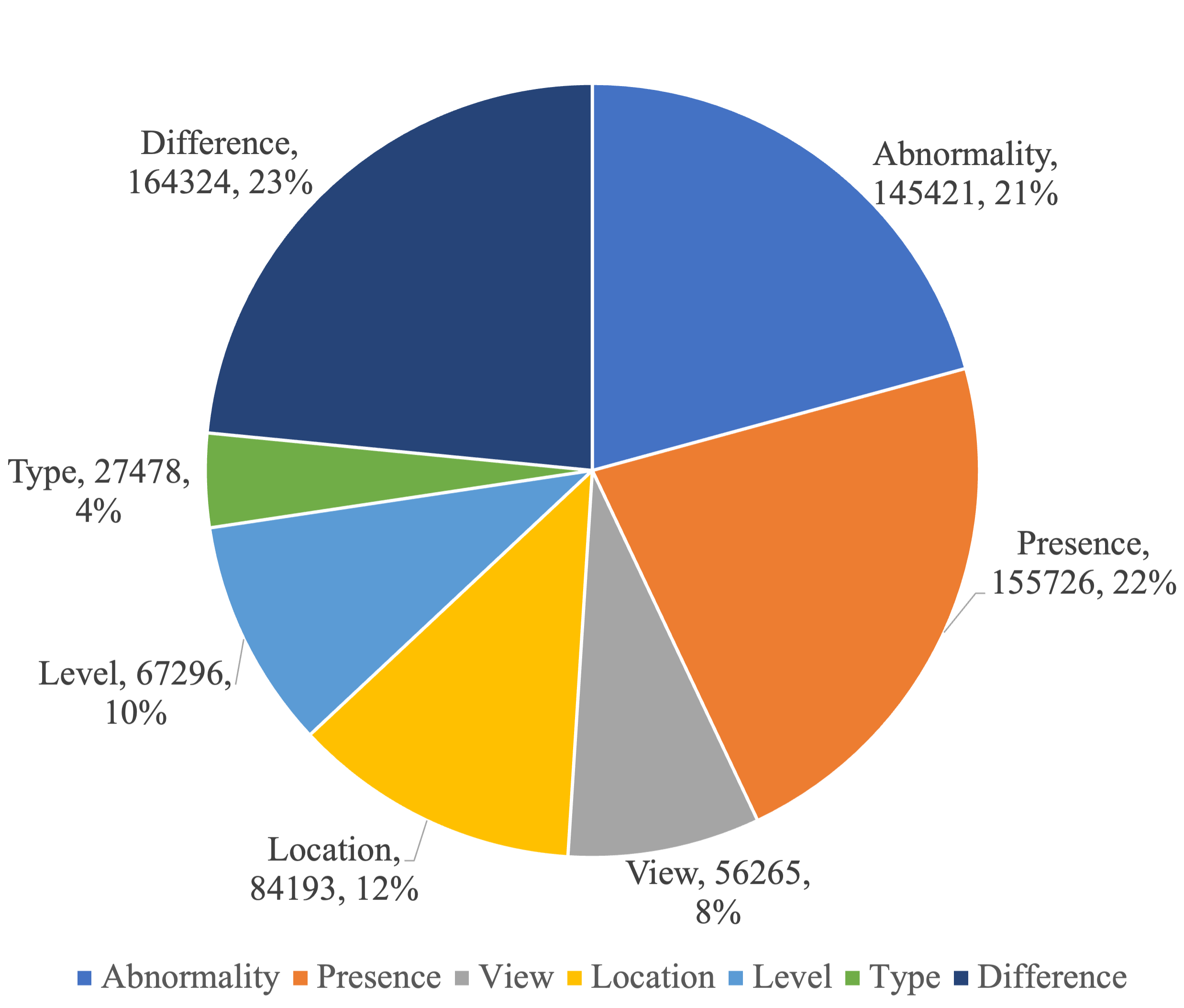}
    \caption{Statistics by question types}
    \Description{A pie chart displaying the number of all the question types.}
    \label{fig:statistics}
\end{figure}

The image pairs are selected from the MIMIC-CXR~\cite{johnson2019mimic} dataset, and each image in an image pair is from the same patient.
A total of \textcolor{black}{164,324} image pairs are selected from MIMIC-CXR, on which \textcolor{black}{700,703} questions are constructed.
We also balance the "yes" and "no" answers to avoid possible bias.
The statistics regarding each question type can be seen in Fig.~\ref{fig:statistics}.
The ratio between the training, validation and testing set is 8:1:1.

\subsection{MIMIC-Diff-VQA dataset construction}
\label{sec:dataset_construct}

% First, we exclude the lateral view and select only the common PA or AP views for comparison.
\textcolor{black}{To ensure the availability of a second image for differential comparison, we excluded patients with only one radiology visit before constructing our dataset. The overall process of dataset construction involves three steps: collecting keywords, building the Intermediate KeyInfo dataset, and generating questions and answers.
}

\subsubsection{\textcolor{black}{Collecting keywords}}
\textcolor{black}{We follow an iterative approach to collect abnormality names and sets of important attributes, such as location, level, and type, from the  MIMIC-CXR dataset. We utilize ScispaCy~\cite{neumann2019scispacy}, a SpaCy model for biomedical text processing, to extract entities from random reports. Subsequently, we manually review all the extracted entities to identify common, frequently occurring keywords that \textcolor{black}{align with radiologists' interests} and add these to our lists of abnormality names and attribute words. We also record different variants of the same abnormality during this process.
The full lists of the selected abnormality names and the attribute words are available in Appendix.}

% We collect a set of abnormality names, as well as the sets of important attributes including location, level, and type, from the MIMIC-CXR dataset.
% The lists of abnormality names and the attribute words are collected by iteratively extracting entities from random reports using ScispaCy~\cite{neumann2019scispacy}, which is a SpaCy model for biomedical text processing. Then we manually go through all the extracted entities that haven't been added to the collection list and select the common keywords that appear frequently \textcolor{black}{and align with radiologists' interests}. Then we add these selected keywords to the collection lists of abnormality names and attributes. 
% During this process, different variants that represent the same abnormality are also recorded.
\subsubsection{\textcolor{black}{Intermediate KeyInfo dataset}}
The previous rule-based label extraction method was limited to a small set of disease-related labels, lacked important information such as complicated disease pathologies, levels, and location, and was prone to errors due to negations. To address these issues, we followed an Extract-Check-Fix cycle to customize the rule set for MIMIC-CXR, ensuring the quality of our dataset through extensive manual verification.

\textcolor{black}{
For each patient visit, we used regular expression rules to extract the abnormality names and their variants. Then, we detected attribute words near the identified abnormalities using these rules. Additionally, by going through the extracted entities, we manually selected the keywords/expressions that indicated negation information to locate the negative findings, i.e. cases where the abnormality did not exist. 
}

% For each study, we use regular expressions to extract the abnormality names as well as their variants then detect attribute words near these detected abnormalities. 
% (Here, "study" represents a single patient visit. Please refer to Section~\ref{datasets} for more context.)
% Meanwhile, by going through the extracted entities, we manually select the keywords/expressions that indicate negation information to localize the negative findings, i.e. cases where the abnormality does not exist.

\textcolor{black}{Next, to ensure the accuracy and completeness of the extracted information, we conducted both manual and automated checks using tools such as Part-of-Speech, ScispaCy entity detection, and MIMIC-CXR-JPG~\cite{johnson2019mimic-jpg} labels as references. These were used to identify any missing or potentially incorrect information that may have been extracted and refined the rules accordingly. We repeated the Extract-Check-Fix cycle until minimal errors were found.}

\begin{figure}[h]
    \centering
    \includegraphics[width=0.45\textwidth]{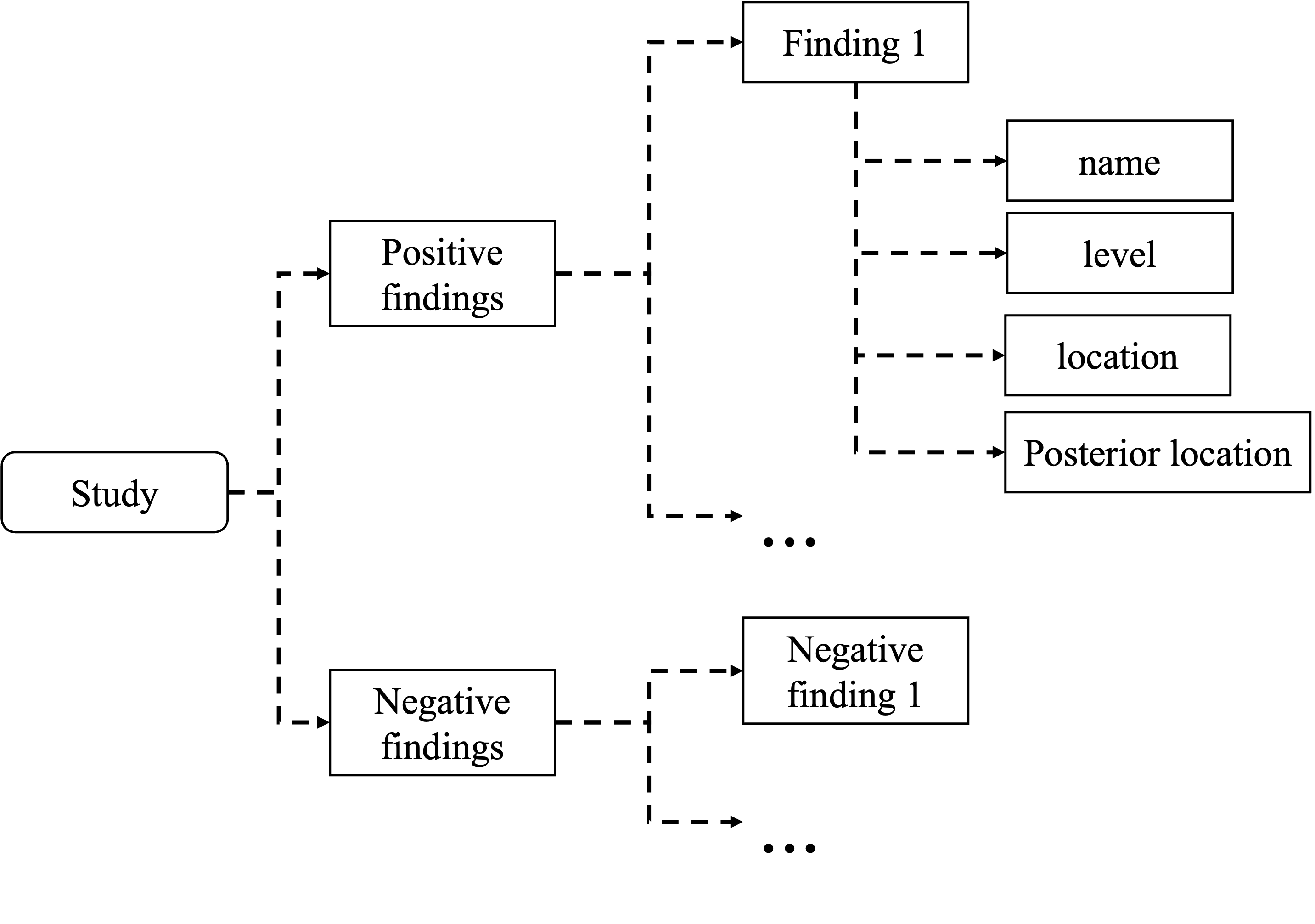}
    \caption{Structure of one study in the Key-Info dataset.}
    \Description{A few boxes connecting to each other, demonstrating the structure of the Key-Info dataset.}
    \label{fig:data_struct}
\end{figure}

\textcolor{black}{
As a result, we have created the Key-Info dataset, consisting of individual study details. 
As shown in Fig.~\ref{fig:data_struct}, for each study, the Key-Info dataset includes information on all positive findings, their attributes, and negative findings.
The "posterior location" attribute represents the location information that appears after the abnormality keyword in a sentence.
}

\subsubsection{Study pairing and question generation}
Once the intermediate Key-Info database is constructed, we can generate study pair questions accordingly.  The examples of each question type are shown in Tab.~\ref{tab:question}. 
Each image pair contains the main image and a reference image, which are extracted from different studies of the same patient. \textcolor{black}{The reference and main visits are chosen strictly based on the earlier visit as the "reference" and the later visit as the "main" image.} Among all the question types, the first six question types are for the main image only, and the \textit{difference} question is for both images.

\subsection{Dataset Validation}
{\color{black}
To further verify the reliability of our constructed dataset, 3 human verifiers were assigned 1700 random sampled question-answer pairs along with the reports and evaluated each sample by annotating "correct" or "incorrect". Finally, the correctness rate of the evaluation achieved 97.33\%, which is acceptable for training neural networks.
Tab.~\ref{tab:evaluation} shows the evaluation results of each verifier.}
It proves that our approach of constructing a dataset in an Extract-Check-Fix cycle works well in ensuring that the constructed dataset has minimum mistakes.

\begin{table}[h]
\color{black}
\caption{\color{black}Evaluation results by human verifiers}
\label{tab:evaluation}
\begin{center}
\begin{tabular}{llll}
\toprule
\multicolumn{1}{c}{\bf Verifier} & \multicolumn{1}{c}{\bf \# of examples} & \multicolumn{1}{c}{\bf \# of correctness} & \multicolumn{1}{c}{\bf Correctness rate} \\
\midrule 
Verifier 1                & 500                              & 475                                 & 95\%                   \\
Verifier 2                & 1000                             & 989                                 & 98.9\%                  \\
Verifier 3                & 200                              & 193                                 & 96.5\%                   \\ 
Total                     & 1700                             & 1657                                 & 97.4\%                  \\
\bottomrule
\end{tabular}
\end{center}
\end{table}

\section{Method}
\label{sec:method}
\subsection{Problem Statement}
Given an image pair $(\I_m, \I_r)$, consisting of the main image $\I_m$ and the reference image $\I_r$, and  a question $\q$, our goal is to obtain the answer $\a$ of the question $\q$ from image pair. In our design, the main and reference images are from the same patient. 
% We follow the diagnostic process of radiologists to compare the changes between the main and the reference images from the same patient and ask different types of questions to accurately localize the changes in the patient's two images and find the abnormalities.

\subsection{Anatomical Structure-Aware Graph Construction and Feature Learning }
\begin{figure*}[t]
    \centering
    \includegraphics[width=0.95\textwidth]{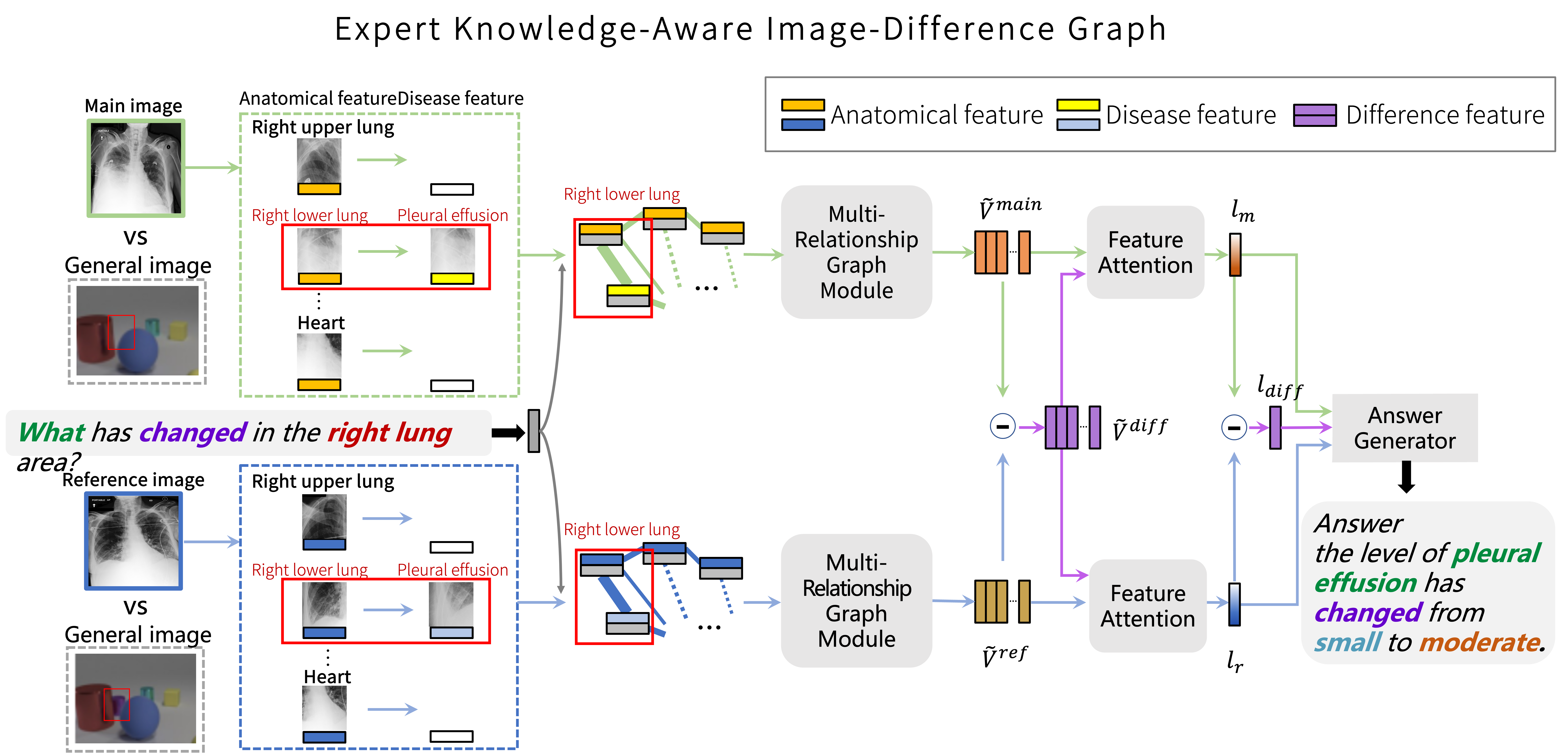}
    \caption{Expert knowledge-aware image-difference graph for medical image difference visual question answering. }
    \Description{The figure shows the architecture of the proposed model. On the left, two chest X-rays are fed into the model, after the anatomical feature disease feature extraction, multi-relationship graph module, feature attention module, and difference feature computation, the final answer is obtained by the Answer Generator.} 
    \label{fig:achitechture}
\end{figure*}

\textcolor{black}{Within the language generation and vision research domain, the most related works to the medical image difference VQA task is image difference captioning \cite{qiu2021describing,oluwasanmi2019fully,yao2022image}, which is designed to identify object movements and changes within a spatial context such as a static or complex background. 
As shown in the left Fig.\ref{fig:achitechture}, the object changes and movements in general image difference captioning are relatively large or significant compared to the background, making the problem easier to solve. These works usually assume a stable background with simple changes in the structure, position, and texture of foreground objects, without significant scaling. 
}
% As shown in the left Fig.\ref{fig:achitechture}, previous work on image difference question answers in the general image domain. They create paired synthetic images with identical backgrounds and only move or remove the simple objects from the background. 
% The feature of image difference was extracted by simply comparing the feature on the exact image coordinates.

\textcolor{black}{
However, the medical image difference is distinct from the general image difference. 
Changes caused by diseases are generally subtle, and the image position, pose, and scale can vary significantly even for the same patient due to the pose and nonrigid deformation. 
}
 \textcolor{black}{As a result, general image difference methods can have difficulty adapting to the medical image difference task.}
To better capture the subtle disease changes and eliminate the pose, orientation, and scale changes, we propose an anatomical structure-aware image difference graph learning solution. Specifically, we represent each anatomical structure as a node and then assess the image changes within each structure in a similar manner to that of radiologists.

\subsubsection{Anatomical Structure,  Disease Region Detection, and Question Encoding.}
To begin, we use a pre-trained Faster-RCNN on the Chest ImaGenome dataset \cite{ren2015faster,wu2021chest, goldberger2000physiobank} to extract the anatomical bounding boxes and their corresponding features $\f_{a}$ from the input images. Subsequently, we train a Faster-RCNN on the VinDr-CXR dataset \cite{pham2021chest} to detect diseases. Rather than directly detecting diseases on the given input images, we extract the features $\f_{d}$ from the same anatomical regions by utilizing the previously extracted anatomical bounding boxes. Following previous work \cite{regat,norcliffe2018learning}, we tokenized each question and answer and embedded them with Glove~(\cite{pennington2014glove}) embeddings. We then used a bidirectional RNN with GRU~\cite{cho2014learning} and self-attention to generate the question embedding $\q$.

% We first extract the anatomical bounding boxes and their features $\f_{a}$ from the input images using pre-trained Faster-RCNN on the MIMIC dataset \cite{ren2015faster,karargyris2020eye}.
% %The anatomical features are extracted in a specific order.
% Then, we train a Faster-RCNN on the VinDr dataset \cite{pham2021chest} to detect the diseases. 
% Instead of directly detecting diseases on the given input images, we extract the features $\f_d$ from the same anatomical regions using the extracted anatomical bounding boxes.
% The questions and answers are processed the same way as ~\cite{regat,norcliffe2018learning}. 
% % After detecting the anatomical and disease regions, we use ROI pooling to extract feature representation on the trained model respectively. 
% Each word is tokenized and embedded with Glove~(\cite{pennington2014glove}) embeddings.
% Then we use a bidirectional RNN with GRU~\cite{cho2014learning} and self-attention to generate the question embedding $\q$.

%We calculated the IOU between the anatomical structure and disease bounding box to determine the location of the diseases. 

%The order of the disease features are assigned according to the anatomical locations.
%By doing so, we ensure each feature from the main image and the reference image corresponds to each other.

%Thereafter, we feed question embedding, the anatomical features and disease features of both main image and reference image to the Multi-Modal Graph Module.

%\textbf{2. Disease Structure Detection and Feature Extraction}
\begin{figure*}[t]
     \centering
     \includegraphics[width=0.88\textwidth]{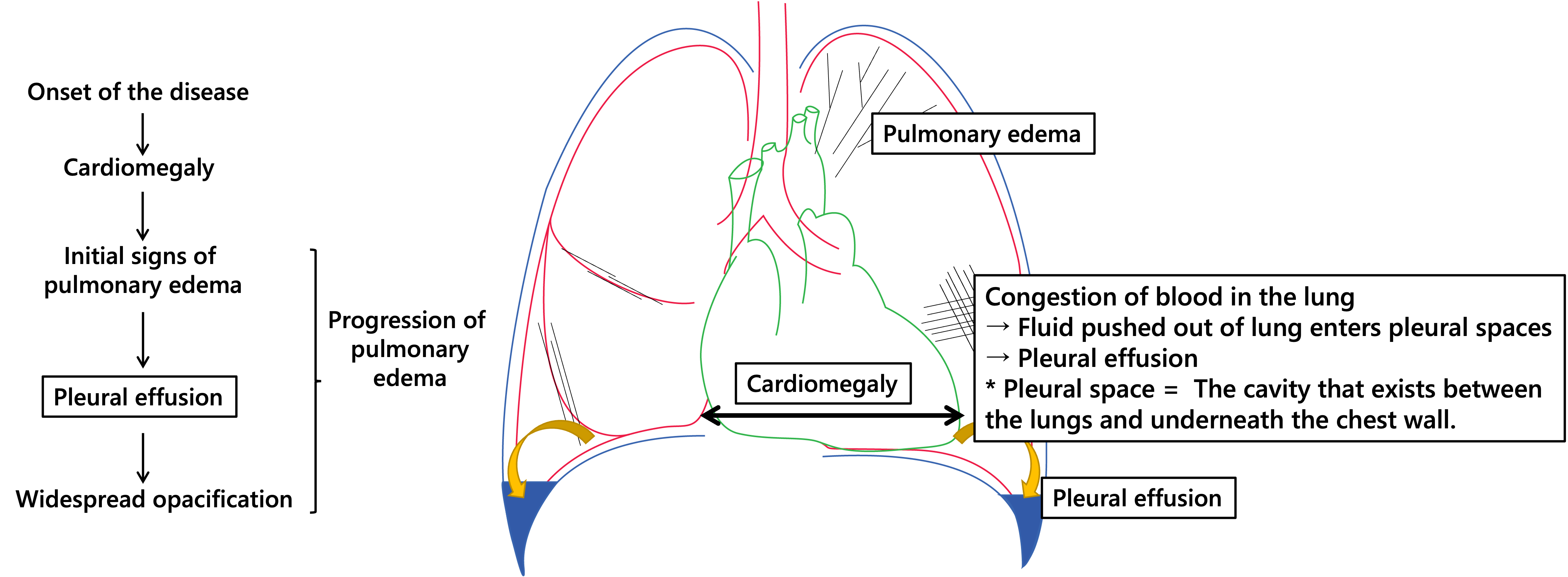}
     \caption{Progression from cardiomegaly to edema and pleural effusion}
     \Description{A sketch of a human thorax, depicting the progression of disease from cardiomegaly to the initial signs of pulmonary edema, to pleural effusion, and finally to widespread opacification.}
     \label{fig:progression}
 \end{figure*}

% \begin{figure*}[t]
%      \begin{subfigure}[b]{\textwidth}
%          \centering
%          \includegraphics[width=0.9\textwidth]{imgs/doctor.png}
%          \caption{Progression from cardiomegaly to edema and pleural effusion}
%          \Description{A sketch of a human thorax, depicting the progression of disease from cardiomegaly to the initial signs of pulmonary edema, to pleural effusion, and finally to widespread opacification.}
%          \label{fig:progression}
%      \end{subfigure}
     
%      % \begin{subfigure}[b]{0.45\textwidth}
%      %     \centering
%      %     \includegraphics[width=\textwidth]{imgs/anno_1.png}
%      %     \caption{Radiologist's annotation example 1}
%      %     \Description{A chest X-ray image with doctor's annotation on the abnormal locations. }
%      %     \label{fig:anno1}
%      % \end{subfigure}
%      % \begin{subfigure}[b]{0.48\textwidth}
%      %     \centering
%      %     \includegraphics[width=\textwidth]{imgs/anno_2.png}
%      %     \caption{Radiologist's annotation example 2}
%      %     \Description{A chest X-ray image with doctor's annotation on the abnormal locations. }
%      %     \label{fig:anno2}
%      % \end{subfigure}
%      \caption{Illustration of the progression of diseases.}
% \end{figure*}

\subsection{Expert Knowledge-Aware Multi-Relationship Graph Module}

\begin{figure}
    \centering
    \includegraphics[width=0.48\textwidth]{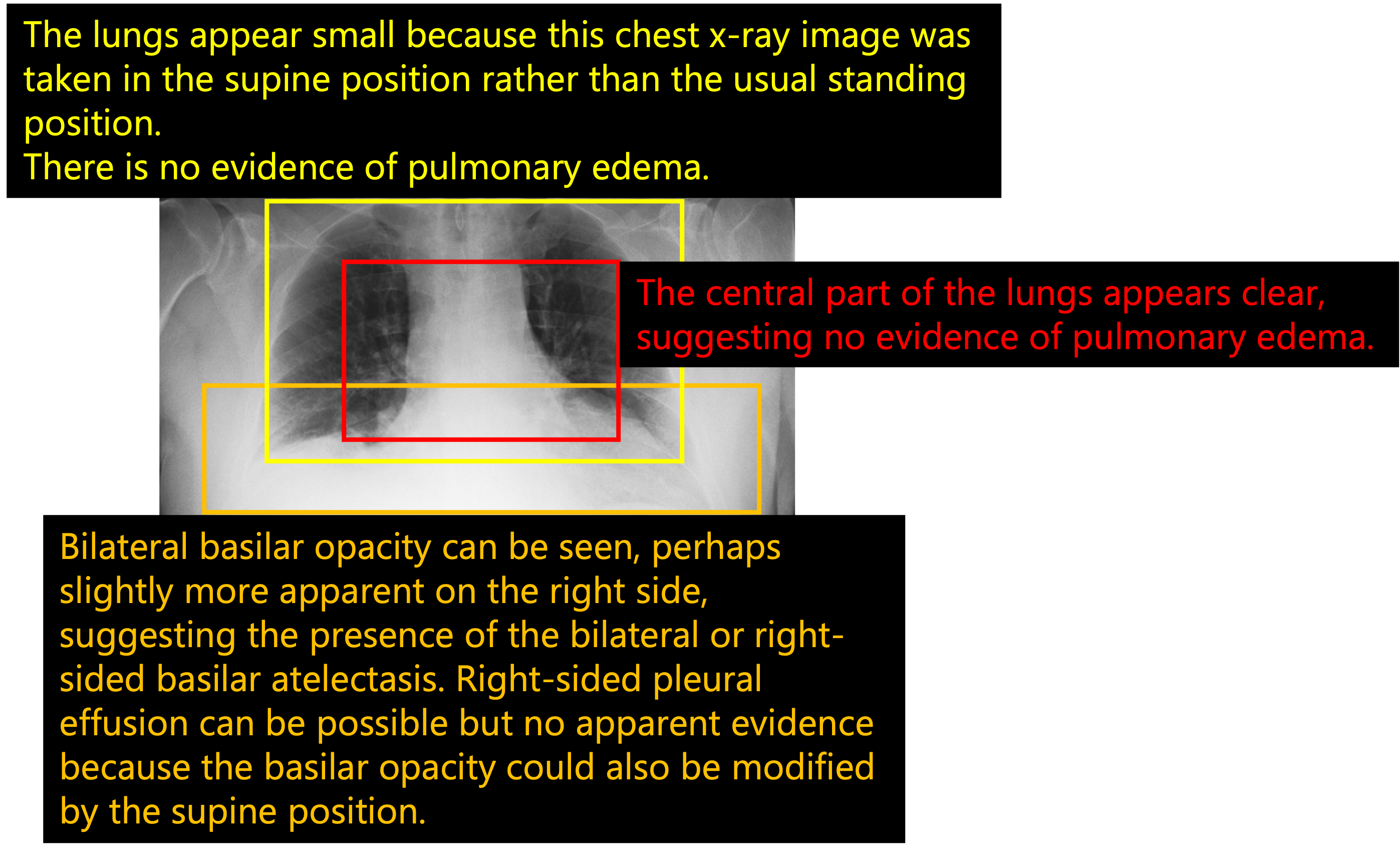}
    \caption{Radiologist's annotation example.}
     \Description{A chest X-ray image with doctor's annotation on the abnormal locations. }
     \label{fig:anno2}
\end{figure}

After extracting the disease and anatomical structure, we construct an expert knowledge-aware image representation graph for the main and reference image. 
The multi-relationship graph is defined as $
\mathcal{G}=\{\V, \mathcal{E}_{sp}, \mathcal{E}_{se}, \mathcal{E}_{imp}\}$, {\color{black}where $\mathcal{E}_{sp}, \mathcal{E}_{se},$ and $ \mathcal{E}_{imp}$ represent the edge sets of spatial graph, semantic graph and implicit graph}, each vertex  $\mathbf{v}_i \in \V, i = 1, \cdots , 2N$ can be either anatomical node~$\mathbf{v}_k = [f_{a,k}\|\q] \in \mathbb{R}^{d_f + d_q}, f_{a,k} \in \f_a, \mathrm{for} \ k = 1, \dots, N$, or disease node~$\mathbf{v}_k=[f_{d,k}\|\q] \in \mathbb{R}^{d_f + d_q}, f_{d,k} \in \f_d,  \mathrm{for} \ k = 1, \dots, N$, representing anatomical structures or disease regions, respectively. 
Both types of nodes are embedded with a question feature as shown in Fig.~\ref{fig:achitechture}. 
$d_f$ is the dimension of the anatomical and disease features.
$d_q$ is the dimension of the question embedding.
$N$ represents the number of anatomical structures of one image. 
Since each disease feature is extracted from the same corresponding anatomical region,
the total number of the vertex is $2N$.

%\textbf{Graph Construction}

We construct three types of relationships in the graph for each image:
1) \textbf{spatial relationship}: 
We construct spatial relationships according to the radiologist's practice of identifying abnormalities based on specific anatomical structures. For example, an actual radiology report can state that "the effusions remain moderate and still
cause substantial bilateral areas of basilar atelectasis" 
% as shown in Fig.~\ref{fig:anno1}; "The central part of the lungs appears clear, suggesting no evidence of pulmonary edema." as shown in Fig.~\ref{fig:anno2}.
In our MIMIC-Diff-VQA dataset, we design questions to assess spatial relationships, such as "Where in the image is the pleural effusion located?" (see Table~\ref{tab:question}). 
Following previous work~\cite{yao2018exploring}, we include 11 types of spatial relations between detected bounding boxes, such as "left lower lung", "right costophrenic angle", etc. The 11 spatial relations includes \texttt{inside} (class1), \texttt{cover} (class2), \texttt{overlap} (class3), and 8 directional classes. Each class corresponds to a 45-degree of direction.
We define the edge between node i and the node j as $a_{ij} = c$, where c is the class of the relationship, $c = 1, 2, \cdots, K$, K is the number of spatial relationship classes, which equals to 11.
When $d_{ij} > t$, we set $\mathbf{a}_{ij}=0$, where $d_{ij}$ is the Euclidean distance between the center points of the bounding boxes corresponding to the node $i$ and node $j$, $t$ is the threshold.
The threshold $t$ is defined as $(l_x + l_y)/3$ by reasoning and imitating the data given by~\cite{regat}.
% When two ROIs are too far, we set their adjacency $\mathbf{a}_{ij}=0 $.  The congested blood can back up into the veins of the lungs.

2) \textbf{Semantic relationship}: 
To incorporate expert knowledge into our approach, we use two knowledge graphs: an anatomical knowledge graph modified from \cite{zhang2020radiology} and a label occurrence knowledge graph built by ourselves. Please refer to the Appendix for detailed information about these knowledge graphs.
% \textcolor{black}{In order to encode expert knowledge,} The semantic relationship is based on two knowledge graphs, including an anatomical knowledge graph modified from \cite{zhang2020radiology}, as shown in Fig.~\ref{fig:anaKG} \textcolor{black}{(the newly added nodes have been marked as red)}, and a label occurrence knowledge graph built by ourselves, as shown in Fig.~\ref{fig:coKG}. 
If two labels are linked by an edge in the knowledge graph, we connect the corresponding nodes in our semantic relationship graph. The knowledge graphs represent abstracted expert knowledge and relationships between diseases, which are essential for disease diagnosis since multiple diseases can interrelate during the progression of a particular disease. For example, Figure~\ref{fig:progression} shows the progression from cardiomegaly to edema and pleural effusion. Cardiomegaly, which refers to an enlarged heart, can result from heart dysfunction that causes blood congestion in the heart, eventually leading to its enlargement. The congested blood is pumped into the lungs' veins, increasing the pressure in the vessels and pushing fluid out of the lungs and into the pleural spaces, indicating the initial sign of pulmonary edema. At the same time, fluid accumulates between the layers of the pleura outside the lungs, resulting in pleural effusion, which can also cause compression atelectasis. If pulmonary edema progresses, widespread opacification will appear in the lungs, as stated in actual diagnostic reports such as "the effusions remain moderate and still cause substantial bilateral areas of basilar atelectasis" and "Bilateral basilar opacity can be seen, suggesting the presence of the bilateral or right-sided basilar atelectasis" (Figure~\ref{fig:anno2}).

% If there is an edge linking two labels in the Knowledge graph, we connect the nodes having these two labels in our semantic relationship graph.
% The knowledge graph represents abstracted expert knowledge and the relationships between diseases. These relationships play a crucial role in disease diagnosis.
% Multiple diseases could be interrelated with each other during the progress of a specific disease.
% For example, in Fig.~\ref{fig:progression}, a progression from cardiomegaly to edema and pleural effusion is shown.
% Cardiomegaly, which refers to an enlarged heart, can start with a heart dysfunction that causes congestion of blood in the heart, eventually leading to the heart's enlargement. The congested blood would be pumped up into the veins of the lungs.
% As the pressure of the vessels in the lungs increases, fluid is pushed out of the lungs and enters pleural spaces, causing the initial sign of pulmonary edema.
% Meanwhile, the fluid starts to build up between the layers of the pleura outside the lungs, \textit{i.e.} pleural effusion.
% Pleural effusion can also cause compression atelectasis.
% If pulmonary edema continues to progress, widespread opacification in the lung will appear.
% We can verify it in actual diagnostic reports. For example, "the effusions remain moderate and still cause substantial bilateral areas of basilar atelectasis"; "Bilateral basilar opacity can be seen, suggesting the presence of the bilateral or right-sided basilar atelectasis" as shown in Fig.~\ref{fig:anno2}.

3) \textbf{Implicit relationship}: \textcolor{black}{a fully connected graph is applied to find the implicit relationships that are not defined by the other two graphs (spatial and semantic graphs). This graph serves as a complement to the other two as it covers all possible relationships, although it is not specific to any one particular relationship.
% $\mathcal{E}_{sp}$, $\mathcal{E}_{se}$ and $\mathcal{E}_{imp}$
% are the set of the spatial, semantic, and implicit edges respectively and $N$ is the number of vertexes.  
Among these three types of relationships, spatial and semantic relationships can be categorized as explicit relationships. The implicit graph itself is categorized as the implicit relationship.}

\subsection{Relation-Aware Graph Attention Network}
\textcolor{black}{
we construct the multi-relationship graph for both main and reference images and use the relation-aware graph attention network (ReGAT) proposed by~\cite{regat} to learn the graph representation for each image. We then embed the image into the final latent feature, which is input into the answer generation module to generate the final answers. Please refer to Appendix for details of the calculation. 
}

\section{Experiments}
\label{exp}
\subsection{Datasets}
\label{datasets}
\textbf{MIMIC-CXR.}
The MIMIC-CXR dataset is a large publicly available dataset of chest radiographs with radiology reports, containing 377,110 images corresponding to 227,835 radiograph studies from 65,379 patients \cite{johnson2019mimic}. One patient may have multiple studies, each consisting of a radiology report and one or more images. Two primary sections of interest in reports are findings: a natural language description of the important aspects of the image, and an impression: a summary of the most immediately relevant findings. 
% Image labels describing findings are derived from either the impression section and the findings section, or the final section of the report \cite{johnson2019mimic-jpg}. Labels are determined by two methods, NegBio \cite{peng2018negbio} and CheXpert \cite{irvin2019chexpert}.
Our MIMIC-Diff-VQA is constructed based on the MIMIC-CXR dataset.

\textbf{Chest ImaGenome.} MIMIC-CXR has been added more annotations by \cite{wu2021chest} including the anatomical structure bounding boxes. This new dataset is named Chest ImaGenome Dataset.
We trained the Faster-RCNN to detect the anatomical structures on their gold standard dataset, which contains 26 anatomical structures.

\textbf{VinDr-CXR.}
The VinDr-CXR dataset consists of 18,000 images manually annotated by 17 experienced radiologists \cite{nguyen2020vindr}. Its images have 22 local labels of boxes surrounding abnormalities and six global labels of suspected diseases. We used it to train the pre-trained disease detection model.

\subsection{Baselines}
\textcolor{black}{It is important to compare multiple baselines. However, we would like to emphasize that the image difference question and answer task is a novel problem even in the general computer vision domain. To date, no prior research has specifically addressed the "image difference question answering" problem. Only a few studies have focused on the general image difference caption task, such as MMCFormers~\cite{qiu2021describing} and IDCPCL~\cite{yao2022image}.  Therefore, our work serves as the first step in this new direction and provides a valuable contribution to the research community.}
\textcolor{black}{We chose baseline models from traditional medical VQA tasks and image difference captioning tasks to address both non-"Difference" and "Difference" queries. Below are the baseline models we have selected:}

% Since we are the first to propose this medical imaging difference VQA problem, we have to choose baseline models from the traditional medical VQA task and image difference captioning task, respectively, \textcolor{black}{to address both non-"Difference" and "Difference" questions. Below are the baseline models we have chosen:}

1.\textit{MMQ} is one of the recently proposed methods to perform the traditional medical VQA task with excellent results. MMQ adopts Model Agnostic Meta-Learning (MAML)~\cite{finn2017model} to handle the problem of the small size of the medical dataset. It also relieves the problem of the difference in visual concepts between general and medical images when finetuning. 

2.\textit{MCCFormers} is proposed to handle the image difference captioning task~\cite{qiu2021describing}. It achieved state-of-the-art performance on the CLEVR-Change dataset~\cite{park2019robust}, a famous image difference captioning dataset. MCCFormers used transformers to capture the region relationships among intra- and inter-image pairs.

{
\color{black}
3.\textit{Image Difference Captioning with Pre-training and Contrastive Learning (IDCPCL)~\cite{yao2022image}} is the state-of-the-art method performed on the general image difference captioning task. They use the pretraining technique to build the bridge between vision and language, allowing them to align large visual variance between image pairs and greatly improve the performance on the challenging image difference dataset, Birds-to-Words~\cite{forbes2019neural}.
}

\subsection{Results and Discussion.}
\label{result}
We implemented the experiments on the PyTorch platform.
We used an Adam optimizer with a learning rate of 0.0001 to train our model for 30,000 iterations at a batch size of 64.
The experiments are conducted on two GeForce RTX 3090 cards with 3 hours and 49 minutes of training time.
The bounding box feature dimension is 1024.
Each word is represented by a 600-dimensional feature vector including a 300-dimensional Glove~\cite{pennington2014glove} embedding.
We used BLEU \cite{papineni2002bleu}, \textcolor{black}{METEOR \cite{lavie2007meteor}, ROUGE\_L \cite{lin2004rouge}, CIDEr \cite{vedantam2015cider}}, which are popular metrics for evaluating the generated text, as the metric in our experiments.
% We used five popular metrics for evaluating the generated text, including BLEU \cite{papineni2002bleu}, METEOR \cite{lavie2007meteor}, ROUGE\_L \cite{lin2004rouge}, CIDEr \cite{vedantam2015cider}, and SPICE \cite{anderson2016spice}.
We obtain the results using Microsoft COCO Caption Evaluation~\cite{chen2015microsoft}.
For the comparison with MMQ, we use accuracy as the metric.

\subsubsection{Ablation Study.}
In Tab.~\ref{tab:metric}, we present the quantitative results of our ablation studies on the MIMIC-Diff-VQA dataset using different graph settings. Our method was tested with implicit graph-only, spatial graph-only, semantic graph-only, and the full model incorporating all three graphs.
As we can see, our full model achieves the best performance across most metrics compared to other graph settings.
Furthermore, in the Appendix, we illustrated the regions of interest (ROIs) of our model using different graphs to demonstrate the improved interpretability achieved by incorporating the spatial and semantic graphs. This is particularly useful in analyzing the location and relationship between abnormalities, providing crucial insights into the anatomical structure from a medical perspective.

\begin{table}[h]
\caption{Quantitative results of our model with different graph settings performed on the MIMIC-Diff-VQA dataset}
\label{tab:metric}
\begin{center}
\begin{tabular}{lllll}
\toprule
\bf Metrics  & \bf Implicit & \bf Spatial & \bf Semantic & \bf Full \\
% \hline\\
\midrule
Bleu-1  & \bf  0.626 & 0.617 & 0.623 & 0.624  \\
Bleu-2  & 0.540 & 0.532 & 0.540 & \bf 0.541  \\
Bleu-3  & 0.475 & 0.468 & \bf 0.477 & \bf 0.477  \\
Bleu-4  & 0.418 &  0.413 & 0.421 & \bf 0.422  \\
METEOR  & 0.333 & 0.337 &\bf  0.340 & 0.337   \\
ROUGE-L & 0.649 &  0.647 & 0.644 &\bf  0.645  \\
CIDEr   & \bf 1.911 &  1.896 & 1.898 & 1.893  \\
% SPICE   & \bf 0.245 & 0.240 & 0.240 & 0.242  \\
\bottomrule
\end{tabular}
\end{center}
\end{table}

% \begin{table}[]
% \centering
% \caption{hhh}
% \begin{tabular}{llllllllll}
% \toprule
% Metric & SPICE & Bleu\_1 & Bleu\_2 & Bleu\_3 & Bleu\_4 & METEOR & ROUGE\_L & CIDEr & SPICE \\
% \midrule
% result & 0.213 & 0.573   & 0.495   & 0.438   & 0.388   & 0.31   & 0.656    & 1.801 & 0.213
% \bottomrule
% \end{tabular}
% \end{table}

\subsubsection{Comparison of accuracy.}
% We also compare our model with the current medical VQA model MMQ on our MIMIC-Diff-VQA dataset.
Due to the nature of MMQ being a classification model, MMQ cannot perform on our \textit{difference} question type because of the diversity of answers.
Also, given that the baseline model cannot take in two images simultaneously, we exclude the \textit{difference} type question from this comparison.
Therefore, we compare our method with MMQ only on the other six types of questions, including \textit{abnormality, presence, view, location, type}, and \textit{level}. These six types of questions have a limited number of answers.
% Since the baseline MMQ is a classification-based model, they used accuracy as their metric.
To compare with them, we use accuracy as the metric for comparison.
Please note that our method is still a text-generation model.
We count the predicted answer as a True answer only when the prediction is fully matched with the ground truth answer.

The comparison results are shown in Tab.~\ref{tab:acc}.
We have refined the comparison into open-ended question results and closed-ended question (with only 'yes' or 'no' answers) results.
It is clear that the current VQA model has difficulty handling our dataset because of the lack of focus on the key regions and the ability to find the relationships between anatomical structures and diseases.
Also, even after filtering out the \textit{difference} questions, there are still \textcolor{black}{9,252} possible answers in total. It is difficult for a classification model to localize the optimal answer from such a huge amount of candidates.

\begin{table}[h]
\caption{Accuracy comparison between our method and MMQ on non-"Difference" questions of the MIMIC-Diff-VQA dataset.}
\label{tab:acc}
\centering
\begin{tabular}{llll}
\toprule
 Question  & Open  & Closed & Total \\
\midrule
MMQ  & 40.5  & 74.2 & 54.7 \\
% MMQ  & 40.5  & 74.2 & 54.7 \\
% Ours & 32.43 & 59.31  & 79.24\\
\bf Ours & 36.6 & 84.9  & 60.2\\
\bottomrule
\end{tabular}
\end{table}

\subsubsection{Evidence and faithfulness}
\textcolor{black}{
In terms of the evidence aspect, our model is designed to enhance the diagnostic process for doctors. Firstly, it highlights the regions of an image indicative of diseases, allowing doctors to quickly and easily inspect and verify their thoughts. Secondly, it empowers doctors to inquire further about specific abnormalities, providing them with the necessary tools to inspect and understand where the information comes from.}

\textcolor{black}{
In terms of the faithfulness aspect, there is concern that the model may capture the distribution of the dataset, relying solely on language priors without comprehending the input image and medical knowledge. To assess this language prior issue, we performed another experiment by removing all images and only keeping the questions. As shown in Tab.~\ref{tab:prior}, the resulting predictions were significantly worse than those obtained using the original images.
}

\begin{table}[h]
\caption{Comparison results between our method using questions only and using both images and questions.}
\label{tab:prior}
\begin{center}
\begin{tabular}{llll}
\toprule
Metrics  & Questions only & Images + questions  \\
% \hline \\
\midrule
Bleu-1   & 0.51     & \bf 0.62 \\
Bleu-2   & 0.33     & \bf 0.54 \\
Bleu-3   & 0.18    & \bf 0.48 \\
Bleu-4   & 0.12     & \bf 0.42 \\
\color{black}METEOR   & \color{black}0.319    & \color{black}\bf 0.337 \\
\color{black}ROUGE\_L & \color{black}0.340      &\color{black} \bf 0.645 \\
\color{black}CIDEr    & \color{black}0          & \color{black}\bf 1.893 \\
% \color{black}SPICE    & -          & \bf 0.457 \\
\bottomrule
\end{tabular}
\end{center}
\end{table}

\subsubsection{Comparison of quality of the text.}
To evaluate the generated answers in the "difference" question, we use metrics specifically designed for evaluating generated text, such as BLEU, METEOR, ROUGE\_L, and CIDEr. The comparison results between our method, MCCFormers, and IDCPCL are presented in Tab.~\ref{tab:mcc}. Our method outperforms MCCFormers in all metrics. Although IDCPCL performs better than MCCFormers, it is still not comparable to our method.

\textcolor{black}{Even though our method utilized the pre-training technique, the comparison is not unfair. The main objective of our pre-trained model is to utilize medical knowledge (read and compare the images in each anatomical structure) to construct graph models and capture subtle changes in images related to disease progression. Our model is specifically tailored for the task of medical image difference VQA and does not employ any general pre-trained strategies like contrastive learning in our framework.
}

\textcolor{black}{
The IDCPCL baseline model used contrastive learning and a combination of three pre-training tasks (Masked Language Modeling, Masked Visual Contrastive Learning, and Fine-grained Difference Aligning) to align images and text. This approach was found to be effective in improving image difference captioning on datasets with large changes and complex background variations. To adapt this approach for the medical image difference VQA task, we made modifications to the IDCPCL model and pre-trained the image and text feature extraction on medical images and clinical notes. Contrastive learning has shown superior performances compared to the conventional pre-trained Resnet classification model~\cite{khosla2020supervised}. Despite the complex pre-training tasks employed, our method significantly outperformed IDCPCL across almost all metrics and interpretability measures.
}

MCCFormers has inferior results compared to our method, as it struggles to differentiate between images. 
This is due to the generated answers of MCCFormers being almost identical and its failure to identify the differences between images. 
MCCFormers, a difference captioning method, compares patch to patch directly, which may work well in the simple CLVER dataset. However, in medical images, most of which are not aligned, the patch-to-patch method cannot accurately identify which region corresponds to a specific anatomical structure. Additionally, MCCFormers does not require medical knowledge graphs to find the relationships between different regions.

% IDCPCL, on the other hand, has the ability to align significant differences between images, enabling it to have higher results than MCCFormers. However, it still uses pre-trained patch-wise image features, which is not feasible in the medical domain where finer features are required.

\begin{table}[h]
\caption{Comparison results between our method and MCCFormers on \textit{difference} questions of the MIMIC-diff-VQA dataset}
\label{tab:mcc}
\begin{center}
\begin{tabular}{llll}
\toprule
\bf Metrics  & \bf  MCCFormers & \bf \color{black}IDCPCL & \bf Ours  \\
% \hline \\
\midrule
Bleu-1   & 0.214  & \color{black}0.614      & \bf 0.628 \\
Bleu-2   & 0.190  & \color{black}0.541     & \bf 0.553 \\
Bleu-3   & 0.170  & \color{black}0.474    & \bf 0.491 \\
Bleu-4   & 0.153  & \color{black}0.414    & \bf 0.434 \\
\color{black}METEOR   & \color{black}0.319    & 0.303      &  \bf  0.339\\
\color{black}ROUGE\_L & \color{black}0.340      & \bf 0.582 &    0.577\\
\color{black}CIDEr    & \color{black}0          & 0.703 &  \bf   1.027\\
\bottomrule
\end{tabular}
\end{center}
\end{table}

\subsubsection{Disccussion}
\textcolor{black}{During the process of clinical reasoning using medical imaging studies, a significant amount of background knowledge is utilized to compare the baseline study (past) with the target study (present). However, modeling background clinical expert knowledge is not straightforward due to its implicitness, which necessitates inferring the best configuration of knowledge modeling based on multiple graphs, such as the implicit, spatial, and semantic graphs (see Figure 3). Therefore, we stand on the shoulder of \cite{regat} which constructs a multi-relationship graph for general image VQA.}

\textcolor{black}{Please note that our model differs fundamentally from the one presented in \cite{regat}. Their model is designed specifically for single-image VQA problems, while ours is for medical image difference VQA, which is a novel problem that involves two images. 
Additionally, our approach extracts anatomical structure-aware features. This involves computing and normalizing the image differences within each anatomical structure, ensuring relevance to disease progression, and invariance to changes in image pose, orientation, and scale. 
To develop our approach, we created an expert knowledge-aware graph that utilizes clinical knowledge. This graph follows the workflow of clinicians who read, compare, and diagnose diseases from medical images based on anatomical structures.
}
\textcolor{black}{Our model is unique in its approach of incorporating clinical knowledge into a multi-relationship graph learning framework, which has not been utilized in general VQA models. 
}

\subsection{Visualization.} Visualized results can be found in Appendix.

\section{Conclusion}
\label{conclusion}
First, We propose a medical image difference VQA problem and collect a large-scale MIMIC-Diff-VQA dataset for this task, which is valuable to both the research and medical communities. Also, we design an anatomical structure-aware \textcolor{black}{feature learning approach and an expert knowledge-aware}  multi-relation image difference graph to extract image-difference features. We train an image difference VQA framework utilizing medical knowledge graphs and compare it to current state-of-the-art methods with improved performances. 
\textcolor{black}{However, there are still limitations to our dataset and method.}
Our constructed dataset currently only focuses on the common cases and ignores special ones, \textit{i.e.} cases where the same disease appears in more than two places. Our current Key-Info dataset can only take care of, at most, two locations of the same disease. 
\textcolor{black}{
Furthermore, there are specific cases where different abnormality names may be combined. For example, when examining edema, interstitial opacities are indicative of edema. Therefore, future work should focus on expanding the dataset to include more special cases.}
% Future work could be extending the dataset to consider more special cases.

\textcolor{black}{
It is worth noting that our model also brings some errors. Representative errors can be summarized into three types: 1, confusion between different presentation aspects of the same abnormality, such as atelectasis and lung opacity being mistaken for each other. 2, different names for the same type of abnormality, such as enlargement of the cardiac silhouette being misclassified as cardiomegaly. 3, the pre-trained backbone (Faster-RCNN) used for extracting image features may provide inaccurate features and lead to incorrect predictions, such as lung opacity being wrongly recognized for pleural effusion.}

\begin{acks}
\textcolor{black}{
This research received support from the JST Moonshot R\&D Grant Number JPMJMS2011 and the Japan Society for the Promotion of Science Grant Number 22K07681. Additionally, it was partially supported by the Intramural Research Program of the National Institutes of Health Clinical Center.
}

\end{acks}

\bibliographystyle{ACM-Reference-Format}
\balance
\bibliography{mybib}

\appendix
\section{Appendix for Visualizations, Related Work, MIMIC-Diff-VQA Dataset, and Our Method}
\label{sec:add_app}
For further information on the Visualizations, related work, MIMIC-Diff-VQA dataset, and our method, please refer to the additional appendix, available at \url{https://github.com/Holipori/KDD2023Appendix/blob/main/Appendix.pdf}.

\end{document}